\documentclass[10pt,twocolumn,letterpaper]{article}

\usepackage{btas}
\usepackage{times}
\usepackage{epsfig}
\usepackage{graphicx}
\usepackage{amsmath}
\usepackage{amssymb}

\usepackage[utf8]{inputenc}
\usepackage{cite}
\usepackage{xcolor}
\usepackage[acronym]{glossaries}
\usepackage[normalem]{ulem}
\usepackage{booktabs}
\usepackage{multirow}
\usepackage{balance}
\usepackage{url}

\usepackage[caption=false]{subfig}

\usepackage[colorlinks=true,pagebackref=false,urlcolor=black,linkcolor=black,citecolor=black]{hyperref} 

\newacronym{cnn}{CNN}{Convolutional Neural Network}
\newacronym{encdec}{ED}{Encoder-Decoder}
\newacronym{fcn}{FCN}{Fully Convolutional Network}
\newacronym{gan}{GAN}{Generative Adversarial Network}
\newacronym{iupui}{IUPUI}{IUPUI Multiwavelength}
\newacronym{masdv1}{MASD.v1}{Multi-Angle Sclera Dataset.v1}
\newacronym{prd}{PRD}{Periocular Region Detection}
\newacronym{ubirisv1}{UBIRIS.v1}{UBIRIS.v1}
\newacronym{ubirisv2}{UBIRIS.v2}{UBIRIS.v2 subset}
\newacronym{ubirisv2s}{UBIRIS.v2}{UBIRIS.v2 subset}
\newacronym{miche}{MICHE-I}{Mobile Iris Challenge Evaluation I}
\newacronym{gs4}{MICHE-GS4}{Galaxy S4 MICHE-I subset}
\newacronym{ip5}{MICHE-IP5}{Apple iPhone 5 MICHE-I subset}
\newacronym{gt2}{MICHE-GT2}{Galaxy Tab 2 MICHE-I subset}
\newacronym{utiris}{UTIRIS}{UTIRIS}
\newacronym{roi}{ROI}{Region of Interest}

\btasfinalcopy 

\ifbtasfinal\fi

\begin{document}
\sloppy

\title{Fully Convolutional Networks and Generative Adversarial \\Networks Applied to Sclera Segmentation}


\author{Diego R. Lucio$^1$, Rayson Laroca$^1$, Evair Severo$^1$, Alceu S. Britto Jr.$^2$, David Menotti$^1$\\
$^1$Federal University of Paran\'a (UFPR), Curitiba, PR, Brazil\\
$^2$Pontifical Catholic University of Paran\'a (PUCPR), Curitiba, PR, Brazil\\
{\tt\small \{drlucio,rblsantos,ebsevero,menotti\}@inf.ufpr.br alceu@ppgia.pucpr.br}
}

\maketitle
\thispagestyle{empty}

\begin{abstract}
Due to the world's demand for security systems, biometrics can be seen as an important topic of research in computer vision. 
One of the biometric forms that has been gaining attention is the recognition based on sclera.
The initial and paramount step for performing this type of recognition is the segmentation of the region of interest, i.e. the sclera.
In this context, two approaches for such task based on the \gls*{fcn} and on \gls*{gan} are introduced in this work. 
\acrshort*{fcn} is similar to a common convolution neural network, however the fully connected layers (i.e., the classification layers) are removed from the end of the network and the output is generated by combining the output of pooling layers from different convolutional ones.
The \acrshort*{gan} is based on the game theory, where we have two networks competing with each other to generate the best segmentation.
In order to perform fair comparison with baselines and quantitative and objective evaluations of the proposed approaches, we provide to the scientific community new $1$,$300$ manually segmented images from two databases\footnote{The new sclera segmentation annotations are publicly available to the research community at \url{https://web.inf.ufpr.br/vri/databases/sclera-segmentation-annotations/}.}.
The experiments are performed on the UBIRIS.v2 and MICHE databases and the best performing configurations of our propositions achieved $F$-$score's$ measures of $87.48\%$ and $88.32\%$, respectively.
\end{abstract}

\vspace{-2mm}
\section{Introduction}

\glsresetall

In recent years, the interest in using biometrics to automatically identify and/or verify a person's identity has greatly increased.
Many characteristics can be used to identify a person, such as physical, biological and behavioral traits~\cite{das2014sclera,zhou2012new}. 
Biometrics are especially important as they can not be changed, forgotten, lost or stolen, providing an unquestionable connection between the individual and the application that makes use of it~\cite{bolle2004guide}.

Several characteristics of the human body can be used as biometrics, such as fingerprint, face, iris, retina and voice, each one with its advantages and disadvantages.
The iris and retina are among the most accurate biometrics~\cite{das2014sclera}.
However, biometric systems  based on iris and retina having high degree of reliability require, respectively, user collaboration and an intrusive image acquisition scheme~\cite{das2015ssbc}. 
In addition to the iris and retina biometric traits, the eye has also a white region around the eyeball known as sclera that contains a pattern of blood vessels that can be used for personal identification~\cite{das2016ssrbc, das2017sserbc, delna2016sclera}.

Typically, segmentation is the first step in which efforts should be applied in a reliable sclera-based recognition system. 
Incorrect segmentation can either reduce the region of blood vessels or introduce new patterns such as eyelashes and eyelids, impairing the system effectiveness.

In order to avoid the above mentioned problems and to encourage the creation of new sclera segmentation techniques, some competitions were performed~\cite{das2015ssbc, das2016ssrbc, das2017sserbc}. 
The state of the art method in these competitions was obtained using a neural network technique named autoencoder in the \gls*{masdv1} database. 
When images of a single sensor were used, the best attained recall and precision rates were $96.65$\% and $95.64$\%, respectively.

In  this work, we proposed two new approaches to sclera segmentation based on \glspl*{cnn}, one based on \gls*{fcn}~\cite{teichmann2016multinet} and another one based on \gls*{gan}~\cite{isola2017imagetoimage}.
To the best of our knowledge, both kinds of networks have never been studied in the sclera segmentation scenario.
\gls*{fcn} is used for segmentation in a large range of applications, from medical to satellite image analysis~\cite{DBLP:journals/corr/abs-1802-01445, Roth2018}, while \gls*{gan} is a new approach to semantic segmentation, which has outperformed the state of the art~\cite{DBLP:journals/corr/LucCCV16}.
The results yielded by the proposed approaches outperform the ones of previous works on a subject of UBIRIS.v2 database~\cite{proenca2010ubirisv2}. 
This subject contains 201 images kindly provided by the authors of~\cite{pinheiro2018progress} and more 300 ones manually labeled by us.
We also present promising results for sclera segmentation on a subset (1000 images) of the \gls*{miche} database~\cite{demarsico_2015_MICHE_database}, which was never been used in this context and initially proposed for iris segmentation and recognition from mobile images.
For producing a fair and quantitative comparison among proposed approaches and a baseline one (SegNet~\cite{badrinarayanan2017segnet}), 
we manually labeled 1,300 images from those databases,  making the masks publicly available for research purposes.
Regarding the \gls*{masdv1} database which was the focus of the previous mentioned competitions for sclera segmentation, we do not have yet implementations on Matlab as required by the organizers of those competitions.

The main contributions of this paper can be summarized as follows:
1) Two new approaches for sclera segmentation;
2) A comparative evaluation of the proposed approaches with a baseline one; 
3) Two datasets composed of $1$,$300$ sclera images manually labeled, being $1$,$000$ from the \gls*{miche} database and $300$ from the \acrshort*{ubirisv2} database.

The remainder of this paper is organized as follows: we briefly review the related work in Section~\ref{sec:related_work}.
In Section~\ref{sec:proposed_approach}, the proposed segmentation approach is described. 
Section \ref{sec:experiments} and~\ref{sec:results} present the experiments and the results obtained, respectively. 
Finally, conclusions and future work are discussed in Section~\ref{conclusions}.
\section{Related Work}
\label{sec:related_work}


In this section, we present a review of the most relevant studies in the sclera segmentation context. We start by presenting some relevant sclera recognition methods in which different sclera segmentation techniques were proposed. Finally, we describe some important works specially dedicated to sclera segmentation techniques. 


\subsection{Sclera-based Recognition Methods}

The works presented in this subsection employed sclera segmentation as preprocessing for sclera-based recognition. Note that, in such cases, the authors did not report the [precision, recall, F-score] achieved in the segmentation stage.

Zhou et al.~\cite{zhou2012new} proposed a new concept for human identification based on the pattern of sclera vessels. In their work, they presented a fully automated sclera segmentation approach for both color and grayscale images. In color images, the sclera region is estimated using the best representation between two color-based techniques. On the other hand, the Otsu's thresholding method is applied to find the sclera region in grayscale images. The \acrshort*{ubirisv1}~\cite{proenca2005ubiris} and \acrshort*{iupui}~\cite{zhou2012new}   databases were used in the experiments.

Das et al.~\cite{das2014sclera} presented a methodology where the right and left sclera are segmented separately. The time adaptive active contour-based region growing segmentation technique proposed by Chan \& Vese~\cite{chan2001active} was employed. The authors applied the Daugman's integro-differential operator to find the seed point required in the region growing-based segmentation \cite{244676}. The \acrshort*{ubirisv1} database was used in the experiments. 

Delna et al.~\cite{delna2016sclera} presented a sclera identification based on a single-board computer. The sclera region is segmented as a rectangle from the Hough circular transform applied for iris location. All images used in the experiments were obtained from a webcam connected to a Raspberry Pi.


\subsection{Sclera Segmentation Techniques}

Das et al.~\cite{das2015ssbc} presented a benchmark for sclera segmentation where four research groups proposed their solutions for this task. All approaches were evaluated in the \gls*{masdv1} database, proposed in the competition. The best results were obtained by Team 4~\cite{radu2015robust}, where the authors presented a novel sclera segmentation algorithm for color images which operates at pixel-level. Exploring various color spaces, the proposed approach is robust to image noise and different gaze directions. The algorithm's robustness is enhanced by a two-stage classifier. At the first stage, a set of simple classifiers is employed, while at the second stage, a neural network classifier operates on the probabilities space generated by the classifier at the first stage. The reported precision and recall rates were $95.05$\% and $94.56$\%, respectively, on the \gls*{masdv1} database.

Das et al.~\cite{das2016ssrbc} proposed a new benchmark, which addresses sclera segmentation and recognition. The best segmentation results reached $85.21$\% precision and $80.21$\% recall. This result is achieved using a method based on Fuzzy C Means, which considers spatial information and uses Gaussian kernel function to calculate the distance  between the center of the cluster and the data points.

Alkassar et al.~\cite{alkassar2016enhanced} proposed a segmentation algorithm which fuses multiple color space skin classifiers to overcome noise introduced by sclera acquisition, such as motion, blur, gaze and rotation. This approach was evaluated using the \acrshort*{ubirisv1}, \acrshort*{ubirisv2}~\cite{proenca2010ubirisv2} and \acrshort*{utiris}~\cite{hosseini2010pigment} databases.

Das et al.~\cite{das2017sserbc} presented a new sclera segmentation benchmark where seven research groups proposed their solutions for this task. The best results attained precision and recall rates of $95.34$\% and $96.65$\%, respectively. 
These results were obtained by using a neural network architecture based on the encoder-decoder approach called SegNet \cite{badrinarayanan2017segnet}.

\subsection{Final Remarks}

As one may see, the use of \glspl*{cnn} was not much explored in the challenging sclera segmentation task. Thus, one of the contributions of this paper is to explore this aspect in the attempt of obtaining improvements in terms of sclera segmentation accuracy.
\section{Proposed Approach}
\label{sec:proposed_approach}

Some images might present specular highlights in regions of the subject's face. In our preliminary tests, many of these regions were erroneously classified as sclera. Therefore, we propose to first locate the periocular region and then perform the sclera segmentation in the detected patch.

This section describes the proposed approach and it is divided into two subsections, one for \gls*{prd} and one for sclera segmentation.

\subsection{\acrlong*{prd}}
\label{sub_sec:eye_detection}

YOLO~\cite{redmon2016yolo} is a object detection framework based on \gls*{cnn}, which regards detection as a regression problem.
As great advances were recently achieved through YOLO-inspired models\cite{laroca2018robust,ning2017spatially}, we decided to fine-tune it for \gls*{prd}. However, as we want to detect only one class and the computational cost is one of our main concerns, we chose to use a smaller model, called Fast-YOLO\footnote{For training Fast-YOLO we used the weights pre-trained on ImageNet, available at \url{https://pjreddie.com/darknet/yolo/}}~\cite{redmon2016yolo}, which uses fewer convolutional layers than YOLO and fewer filters in those layers. 
The Fast-YOLO’s architecture is shown in Table~\ref{tab:fast_yolo2}.

\begin{table}[!htb]
	\centering	
	\caption{Fast-YOLO network used to detect the periocular region. We reduced the number of filters in the last convolutional layer from $125$ to $30$ in order to output $1$ class instead of $20$.}
	\vspace{1mm}
	\label{tab:fast_yolo2}	
	\resizebox{\columnwidth}{!}{	
		\begin{tabular}{@{}cccccc@{}}
			\toprule
			\multicolumn{2}{c}{\textbf{Layer}} & \textbf{Filters} & \textbf{Size} & \textbf{Input} & \textbf{Output} \\ \midrule
			$0$ & conv & $16$ & $3 \times 3 / 1$ & $416 \times 416 \times 1 / 3$ & $416 \times 416 \times 16$ \\
			$1$ & max &  & $2 \times 2 / 2$ & $416 \times 416 \times 16$ & $208 \times 208 \times 16$ \\
			$2$ & conv & $32$ & $3 \times 3 / 1$ & $208 \times 208 \times 16$ & $208 \times 208 \times 32$ \\
			$3$ & max &  & $2 \times 2 / 2$ & $208 \times 208 \times 32$ & $104 \times 104 \times 32$ \\
			$4$ & conv & $64$ & $3 \times 3 / 1$ & $104 \times 104 \times 32$ & $104 \times 104 \times 64$ \\
			$5$ & max &  & $2 \times 2 / 2$ & $104 \times 104 \times 64$ & $52 \times 52 \times 64$ \\
			$6$ & conv & $128$ & $3 \times 3 / 1$ & $52\times 52 \times 64$ & $52 \times 52 \times 128$ \\
			$7$ & max &  & $2 \times 2 / 2$ & $52 \times 52 \times 128$ & $26 \times 26 \times 128$ \\
			$8$ & conv & $256$ & $3 \times 3 / 1$ & $26 \times 26 \times 128$ & $26 \times 26 \times 256$ \\
			$9$ & max &  & $2 \times 2 / 2$ & $26 \times 26 \times 256$ & $13 \times 13 \times 256$ \\
			$10$ & conv & $512$ & $3 \times 3 / 1$ & $13 \times 13 \times 256$ & $13 \times 13 \times 512$ \\
			$11$ & max &  & $2 \times 2 / 1$ & $13 \times 13 \times 512$ & $13 \times 13 \times 512$ \\
			$12$ & conv & $1024$ & $3 \times 3 / 1$ & $13 \times 13 \times 512$ & $13 \times 13 \times 1024$ \\
			$13$ & conv & $1024$ & $3 \times 3 / 1$ & $13 \times 13 \times 1024$ & $13 \times 13 \times 1024$ \\
			$14$ & conv & $30$ & $1 \times 1 / 1$ & $13 \times 13 \times 1024$ & $13 \times 13 \times 30$ \\
			$15$ & detection &  &  &  &  \\ \bottomrule
		\end{tabular}}
		\vspace*{-6pt}
		
\end{table}

The \gls*{prd} network is trained using the images, without any preprocessing, and the coordinates of the \gls*{roi} as inputs. As ground truth, we used the annotations provided by Severo et al.~\cite{severo2018benchmark}. As these annotations were made for iris location, we applied a padding (chosen based on the validation set) in the detected patch (i.e., iris), so that the sclera is entirely within the \gls*{roi}.

By default, Fast-YOLO only returns objects detected with a confidence of $0.25$ or higher. We consider only the detection with the largest confidence in cases where more than one periocular region is detected, since there is always only one region annotated in the evaluated databases. If no region is detected, the next stage (sclera segmentation) is performed on the image in its original size.

\subsection{Sclera Segmentation}
\label{sub_sec:sclera_segmentation}

We employ three approaches for sclera segmentation, since they presented good results in other segmentation applications. These approaches are: \gls*{fcn}, \gls*{encdec} and \gls*{gan}. Its noteworthy that \gls*{encdec} was employed for sclera segmentation in~\cite{das2017sserbc}, obtaining state-of-the-art results. Therefore, we made use of \gls*{encdec} in the databases used in this paper, considering it as the baseline for comparison with the proposed approach.

\subsubsection{\acrlong*{fcn}}
\label{sub_sub_sec:fully_connected_network}

This segmentation approach was proposed by Long et al.~\cite{long2015fully}. The network has only convolutional layers and the segmentation process can take input images of arbitrary sizes, producing correspondingly-sized output with efficient inference and learning.

\begin{figure}[!htb]
	\begin{center}
		\includegraphics[scale=0.232]{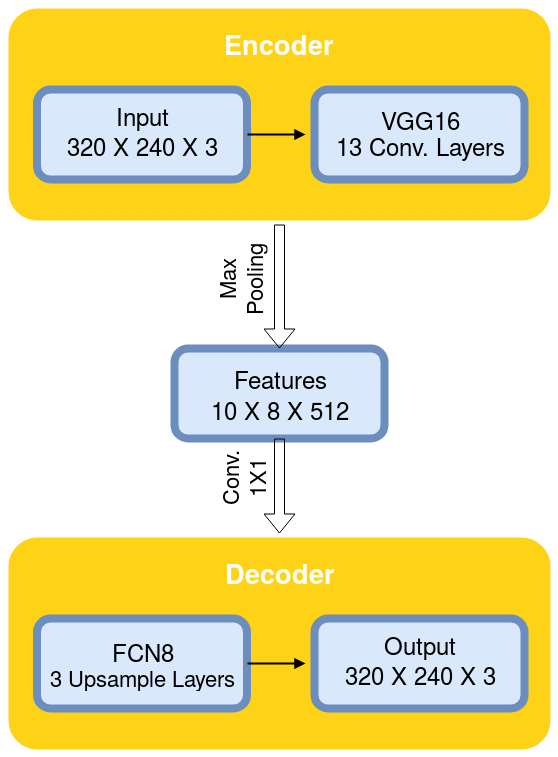}
	\end{center}
	\vspace{-2mm}
	\caption{\gls*{fcn} architecture for sclera segmentation.}
	\label{fig:architecture_fcn}
\end{figure}

In this work, we employ the \gls*{fcn} approach presented by Teichmann et al.~\cite{teichmann2016multinet}. As shown in Figure~\ref{fig:architecture_fcn}, features are extracted using a \gls*{cnn} without the fully connected layers (i.e., VGG-$16$ without the last $3$ layers).

Next, the extracted features pass through two $1\times1$ convolutional layers, generating an output of dimension $10\times8\times 6$. The output of these convolutional layers is processed by the FCN8 architecture proposed in~\cite{long2015fully}, which performs the up-sampling combining the last three layers from the VGG-$16$. 

\subsubsection{\acrlong*{encdec}}
\label{sub_sub_sec:auto_encoder}

\begin{table*}[!htb]
	\centering
	\caption{SegNet architecture.}
	\vspace{1mm}
	\label{tab:encoder}
	\resizebox{\textwidth}{!}{
		\begin{tabular}{@{}cccccc|cccccc@{}}
			\toprule
			\multicolumn{2}{c}{\textbf{Layer}} & \textbf{Filters} & \textbf{Size} & \textbf{Input} & \textbf{Output} & \multicolumn{2}{c}{\textbf{Layer}} & \textbf{Filters} & \textbf{Size} & \textbf{Input} & \textbf{Output}\\ \midrule
			$1$ & enc & 64  & $3 \times 3$ & $320 \times 240 \times 3/1$ & $320 \times 240 \times 64$ & $19$ & up &  & $2 \times 2$ & $10 \times 8 \times 512$ & $20 \times 15 \times 512$ \\ 
			$2$ & enc & 64 & $3 \times 3$ & $320 \times 240 \times 64$ & $320 \times 240 \times 64$ & $20$ & dec & 512 & $3 \times 3$ & $20 \times 15 \times 512$ & $20 \times 15 \times 512$ \\
			$3$ & max &  & $2 \times 2$ &$320 \times 240 \times 64$ & $160 \times 120 \times 64$ & $21$ & dec & 512 & $3 \times 3$ & $20 \times 15 \times 512$ & $20 \times 15 \times 512$\\
			$4$ & enc & 128  & $3 \times 3$ & $160 \times 120 \times 64$ & $160 \times 120 \times 128$ & $22$ & dec & 512 & $3 \times 3$ & $20 \times 15 \times 512$ & $20 \times 15 \times 512$\\
			$5$ & enc & 128 & $3 \times 3$ & $160 \times 120 \times 128$ & $160 \times 120 \times 128$ & $23$ & up &  & $2 \times 2$ & $20 \times 15 \times 512$ & $40 \times 30 \times 512$\\
			$6$ & max &  & $2 \times 2$ & $160 \times 120 \times 128$ & $80 \times 60 \times 128$ & $24$ & dec & 512 & $3 \times 3$ & $40 \times 30 \times 512$ & $40 \times 30 \times 512$\\
			$7$ & enc & 256 & $3 \times 3$ & $80 \times 60 \times 128$ & $80 \times 60 \times 256$ & $25$ & dec & 512 & $3 \times 3$ & $40 \times 30 \times 512$ & $40 \times 30 \times 512$ \\
			$8$ & enc & 256 & $3 \times 3$ & $80 \times 60 \times 256$ & $80 \times 60 \times 256$ & $26$ & dec & 256 & $3 \times 3$ & $40 \times 30 \times 512$ & $40 \times 30 \times 256$\\
			$9$ & enc & 256 & $3 \times 3$ & $80 \times 60 \times 256$ & $80 \times 60 \times 256$ & $27$ & up &  & $2 \times 2$ & $40 \times 30 \times 256$ & $80 \times 60 \times 256$ \\
			$10$ & max &  & $2 \times 2$ & $80 \times 60 \times 256$ & $40 \times 30 \times 256$ & $28$ & dec & 256 & $3 \times 3$ & $80 \times 60 \times 256$ & $80 \times 60 \times 256$ \\
			$11$ & enc & 512 & $3 \times 3$ & $40 \times 30 \times 256$ & $40 \times 30 \times 512$ & $29$ & dec & 256 & $3 \times 3$ & $80 \times 60 \times 256$ & $80 \times 60 \times 256$\\
			$12$ & enc & 512 & $3 \times 3$ & $40 \times 30 \times 512$ & $40 \times 30 \times 512$ & $30$ & dec & 128 & $3 \times 3$ & $80 \times 60 \times 256$ & $80 \times 60 \times 128$\\
			$13$ & enc & 512 & $3 \times 3$ & $40 \times 30 \times 512$ & $40 \times 30 \times 512$ & $31$ & up &  & $2 \times 2$ & $80 \times 60 \times 512$ & $160 \times 120 \times 128$\\
			$14$ & max &  & $2 \times 2$ & $40 \times 30 \times 512$ & $20 \times 15 \times 512$ & $32$ & dec & 128 & $3 \times 3$ & $160 \times 120 \times 128$ & $160 \times 120 \times 128$ \\
			$15$ & enc & 512 & $3 \times 3$ & $20 \times 15 \times 512$ & $20 \times 15 \times 512$ & $33$ & dec & 64 & $3 \times 3$ & $160 \times 120 \times 128$ & $160 \times 120 \times 64$ \\
			$16$ & enc & 512 & $3 \times 3$ & $20 \times 15 \times 512$ & $20 \times 15 \times 512$ & $34$ & up &  & $2 \times 2$ & $160 \times 120 \times 64$ & $320 \times 240 \times 64$\\
			$17$ & enc & 512 & $3 \times 3$ & $20 \times 15 \times 512$ & $20 \times 15 \times 512$ &$35$ & dec & 64 & $3 \times 3$ & $320 \times 240 \times 64$ & $320 \times 240 \times 64$ \\
			$18$ & max &  & $2 \times 2$ & $20 \times 15 \times 512$ & $10 \times 8 \times 512$ &$36$ & dec & 2  & $3 \times 3$ & $320 \times 240 \times 64$ & $320 \times 240 \times 2$\\
						 \bottomrule
		\end{tabular}}
		\vspace{-8pt}
		
	\end{table*}

A convolutional \gls*{encdec}, also called autoencoder, is a neural network trained in order to copy its input to the output. The purpose is to learn data encoding which can be used for dimensionality reduction or even for file compression \cite{audebert2017semantic}.

The \gls*{encdec} (SegNet) used in this work was presented in~\cite{badrinarayanan2017segnet}. SegNet consists of a stack of encoders followed by a corresponding stack of decoders which feed a soft-max classification layer. Decoders map low-resolution features extracted by encoders to an image with the same dimension as the input. The architecture used is presented in Table~\ref{tab:encoder}.
	
\subsubsection{\acrlong*{gan}}
\label{sub_sub_sec:generative_adversarila_network}

\begin{figure}[!htb]
	\begin{center}
		\includegraphics[scale=0.35]{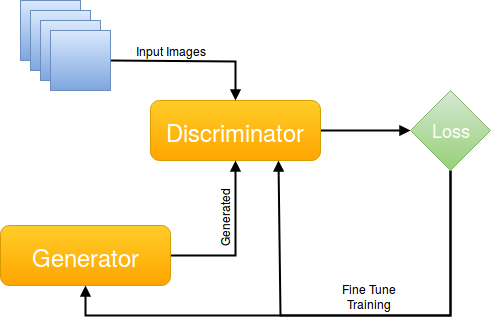}
	\end{center}
	\vspace{-2mm}
	\caption{\gls*{gan} architecture for sclera segmentation.}
	\label{fig:architecture_gan}
\end{figure}

\glspl*{gan} are deep neural networks composed by both generator and discriminator networks, pitting one against the other. First, the generator network receives noise as input and generates samples. Then, the discriminator network receives samples of training data and those of the generator network, being able to distinguish between the two sources~\cite{goodfellow2014generative}. A generic GAN architecture is shown in Figure~\ref{fig:architecture_gan}. 

Basically, the generator network learns to produce more realistic samples throughout each iteration, and the discriminator network learns to better distinguish real and synthetic datas.

Isola et al.~\cite{isola2017imagetoimage} presented the \gls*{gan} approach used in this work, which is a conditional \gls*{gan} able to learn the relation between a image and its label file, and, from that, generate a variety of image types, which can be employed in various tasks such as photo generation and semantic segmentation.
\section{Experiments}
\label{sec:experiments}

In this section, we present the databases and the evaluation protocol used in our experiments.

\subsection{Databases}
\label{sub_sec:databases}

The experiments were carried out in two subsets of well-known iris databases: \acrshort*{ubirisv2} and \acrshort*{miche}. An overview of both subsets can be seen in Table~\ref{tab:overview_datasets}. Remark that we do not use the SSBC~\cite{das2015ssbc} and SSRBC~\cite{das2016ssrbc} databases in our experiments as only the test sets were made available by the authors.

\begin{table}[!htb]
	\centering
	\caption{Overview of the databases used in this work. All of these are a subset of the original database.}
	\vspace{1mm} 
	\label{tab:overview_datasets}
			\begin{tabular}{@{}cccc@{}}
				\toprule
				\textbf{Database}    & \textbf{Images} & \textbf{Subjects} & \textbf{Resolution}  \\ \midrule
				\acrshort*{ubirisv2} & $500$ & $261$ &  $400 \times 300$ \\
				
				\acrshort*{miche} & $1$,$000$ & $92$ & Various      \\ 
				\acrshort*{gs4} & $333$ & $92$ & Various      \\
				\acrshort*{ip5} & $323$ & $92$ & Various      \\
				\acrshort*{gt2} & $344$ & $92$ & $640 \times 480$      \\
				\bottomrule
			\end{tabular}
\end{table}

\noindent
\textbf{\acrshort*{ubirisv2}}: this database is composed of  $11$,$102$ images collected from both eyes from $261$ subjects and have a resolution of $400\times300$ pixels.

\noindent \textbf{\acrshort*{miche}}: this database consists of $3$,$191$ images captured from $92$ subjects under uncontrolled settings using three mobile devices: iPhone~5, Galaxy Samsung~IV and Galaxy Tablet~II ($1$,$262$, $1$,$297$ and $632$ images, respectively), with many different resolutions \cite{demarsico_2015_MICHE_database}. 

\subsection{Preprocessing}
\label{sub_sec:pre_processing}

\begin{figure}[!htb]
	\vspace{-5mm}
	\begin{center}
		\subfloat[][\acrshort*{ip5}]{
			\includegraphics[width=0.24\columnwidth]{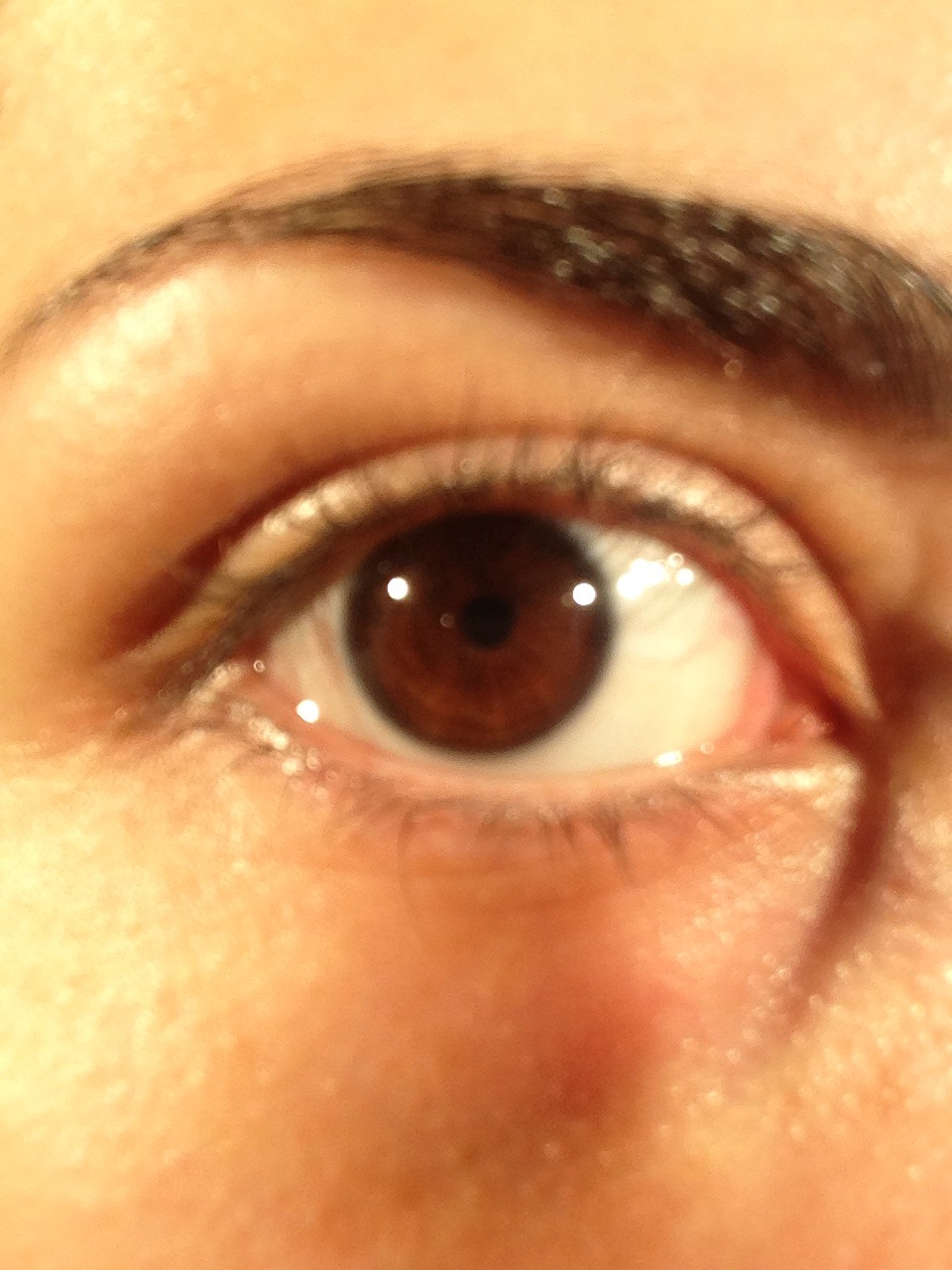}%
			\includegraphics[width=0.24\columnwidth]{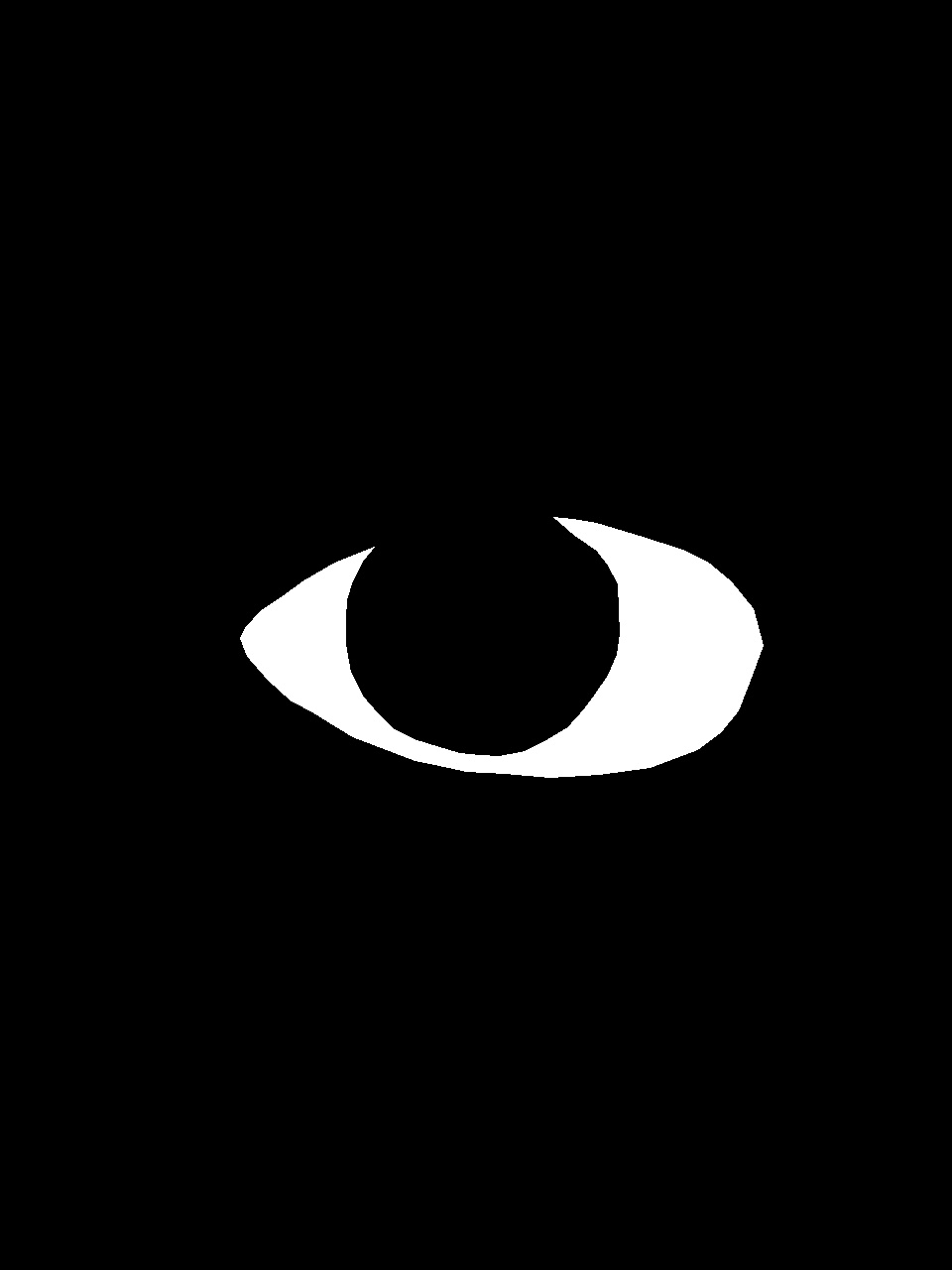}}\hfill
		\subfloat[][\acrshort*{gs4}]{
			\includegraphics[width=0.24\columnwidth]{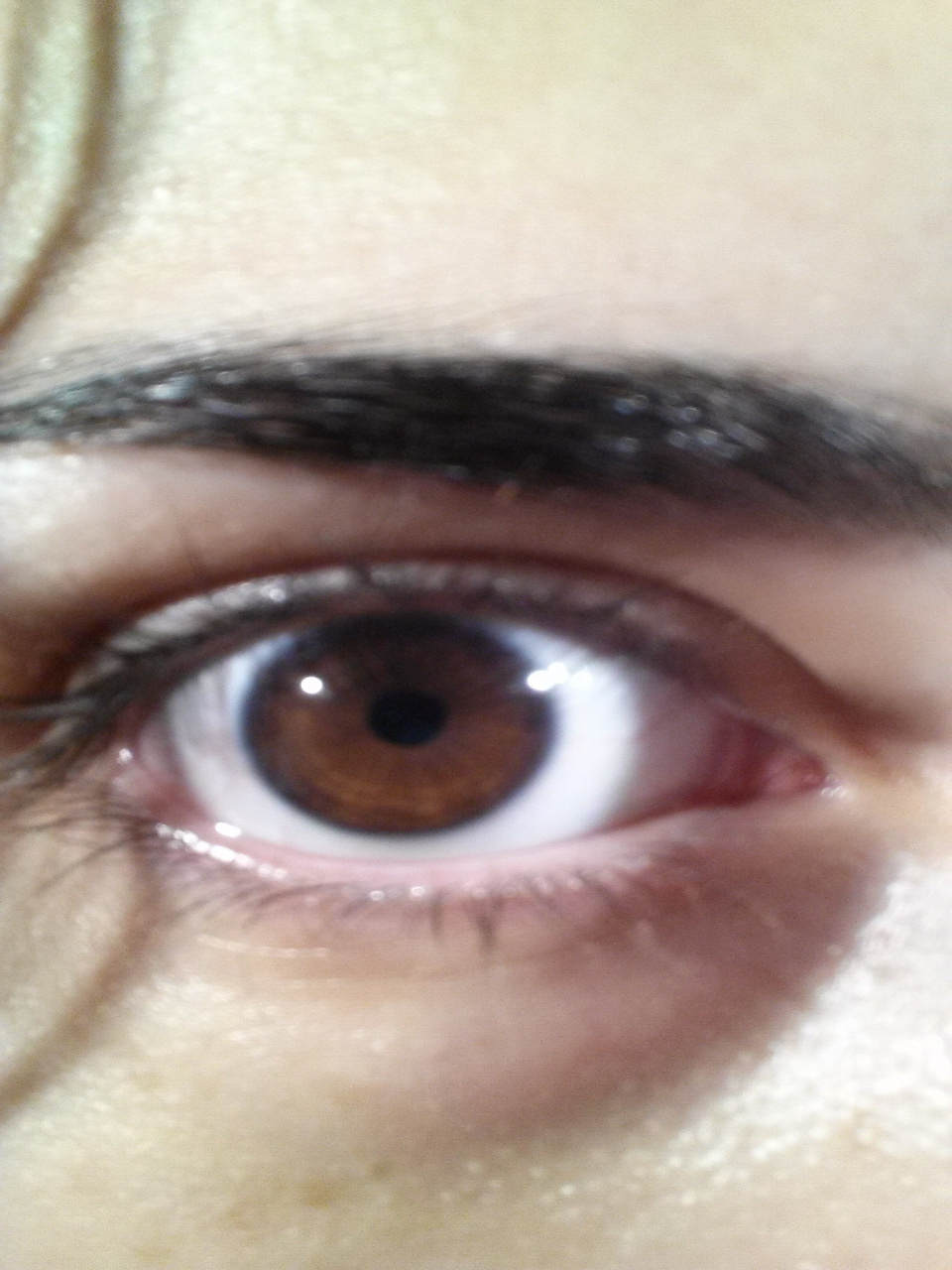}%
			\includegraphics[width=0.24\columnwidth]{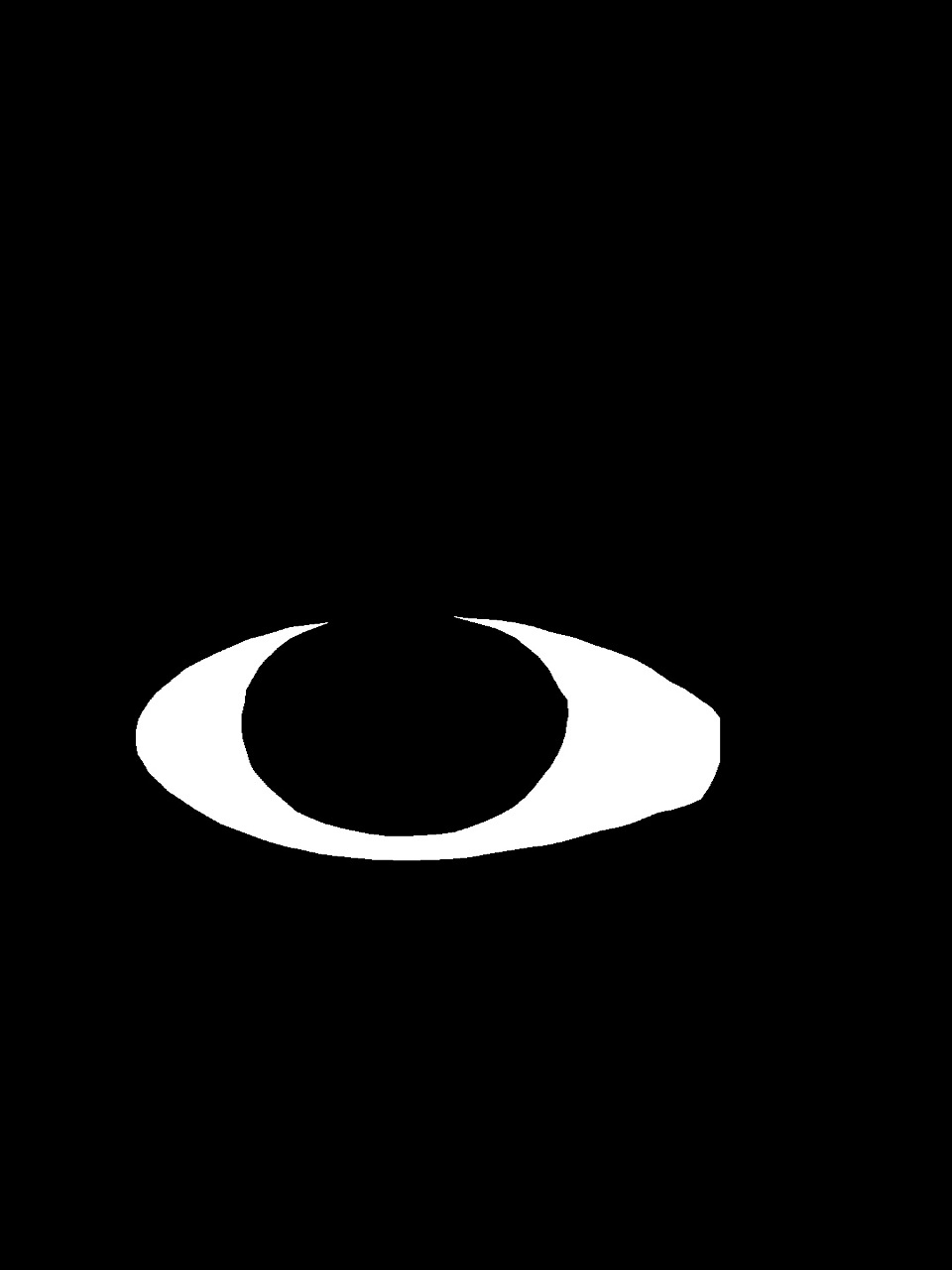}} \\[-1ex]
		
		\subfloat[][\acrshort*{gt2}]{
			\includegraphics[width=0.24\columnwidth]{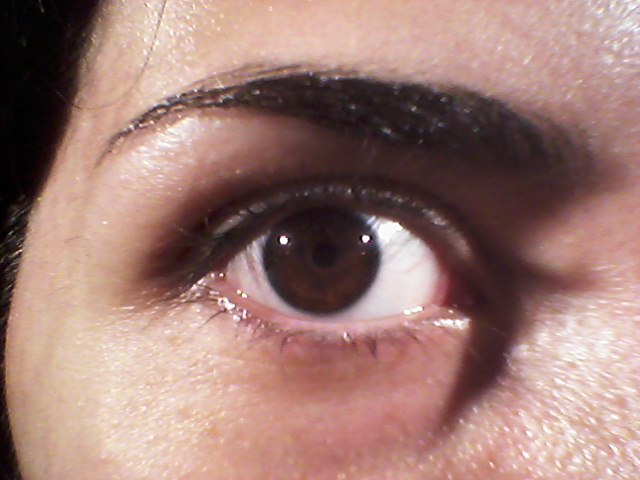}%
			\includegraphics[width=0.24\columnwidth]{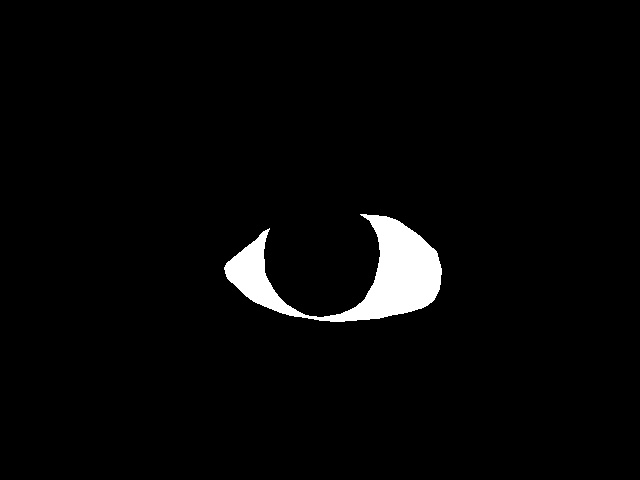}}\hfill
		\subfloat[][\acrshort*{ubirisv2s}]{
			\includegraphics[width=0.24\columnwidth]{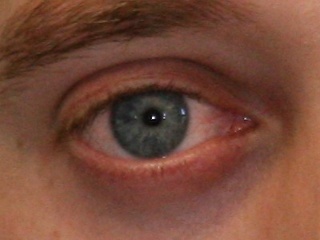}%
			\includegraphics[width=0.24\columnwidth]{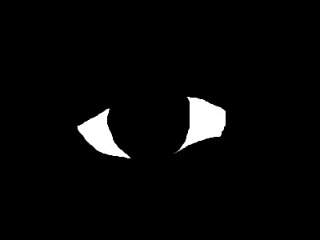}}
	\vspace{1mm}
	\caption{Four examples of the masks created by us.}
	\label{fig:masks_original}
	\end{center}
\end{figure}

As discussed in Section~\ref{sub_sec:eye_detection}, it was necessary to first detect the periocular region as some images present specular highlights, impairing the performance of the system. After detecting the periocular region, only the \gls*{roi} was maintained in each image, providing a great improvement over the results obtained at first. Figure~\ref{fig:masks_original} shows an example of each subset of the original image (without \gls*{prd}) and the segmentation mask created by us. 

\begin{figure}[!htb]
	\vspace{-5mm}
	\begin{center}
		\subfloat[][\acrshort*{ip5}]{
			\includegraphics[width=0.24\columnwidth]{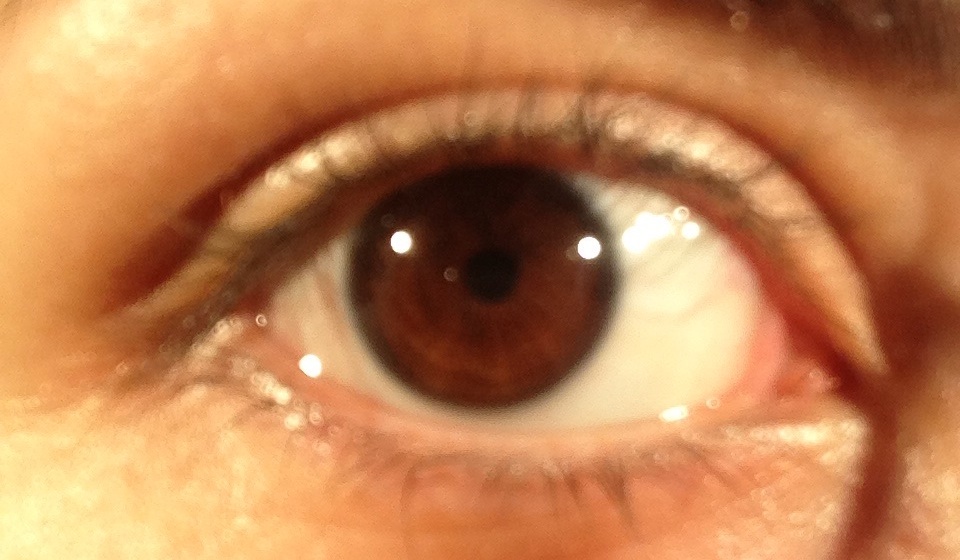}%
			\includegraphics[width=0.24\columnwidth]{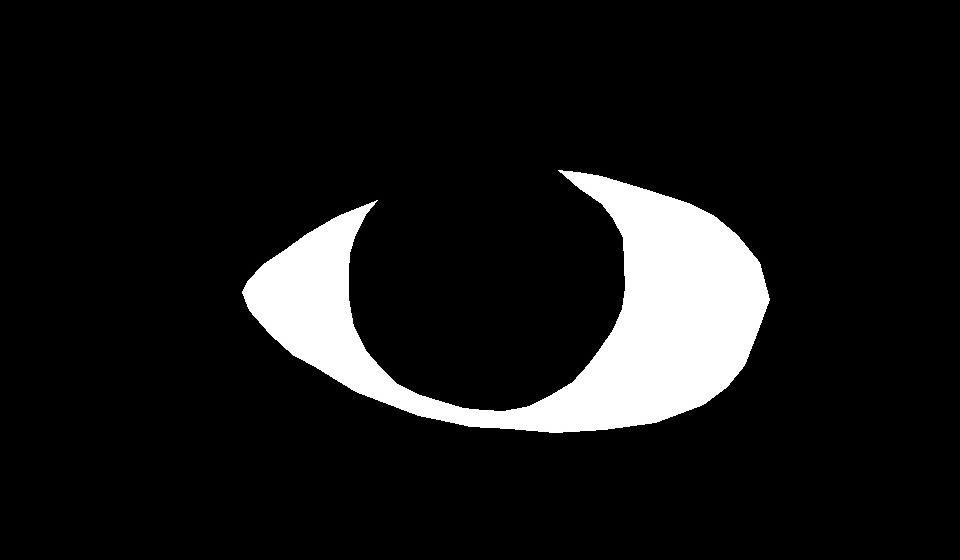}}\hfill
		\subfloat[][\acrshort*{gs4}]{
			\includegraphics[width=0.24\columnwidth]{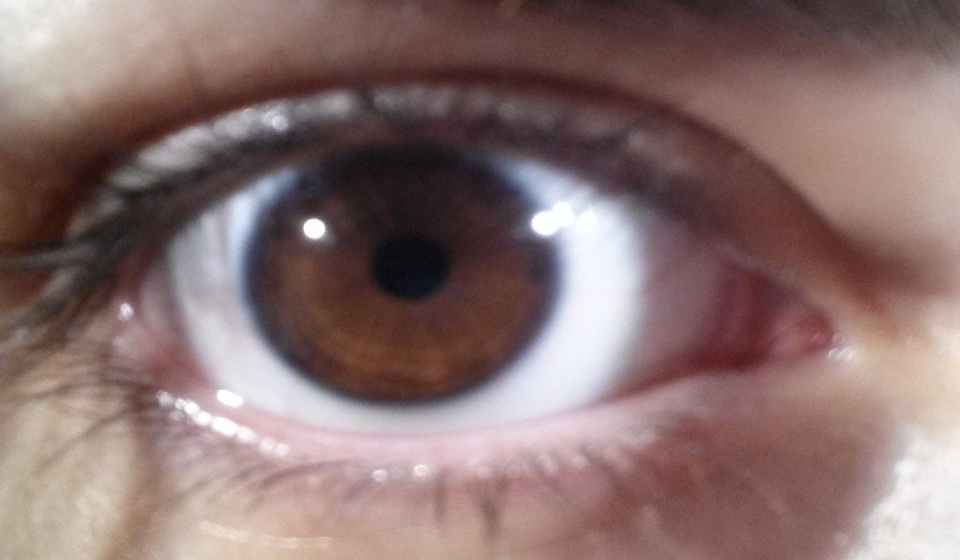}%
			\includegraphics[width=0.24\columnwidth]{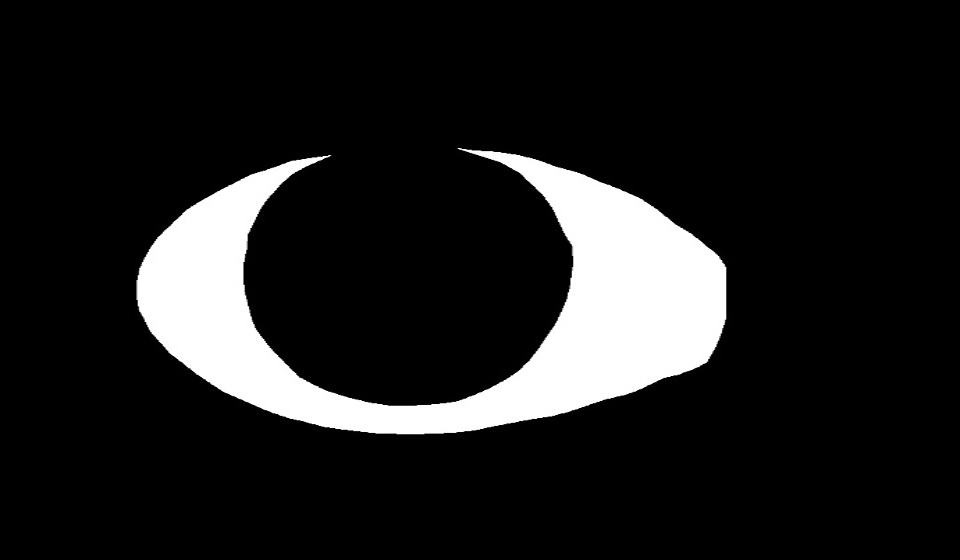}} \\[-1ex]
		
		\subfloat[][\acrshort*{gt2}]{
			\includegraphics[width=0.24\columnwidth]{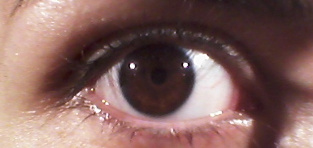}%
			\includegraphics[width=0.24\columnwidth]{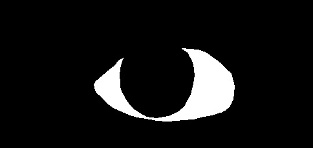}}\hfill
		\subfloat[][\acrshort*{ubirisv2s}]{
			\includegraphics[width=0.24\columnwidth]{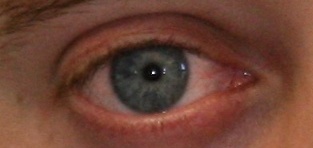}%
			\includegraphics[width=0.24\columnwidth]{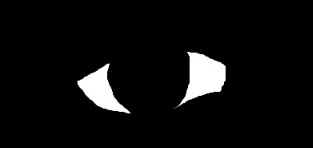}}
			
			\vspace{1mm}
			\caption{Periocular regions detected and replicated to the masks.}
			\label{fig:masks} 
	\end{center}
\end{figure}

\vspace{-4mm}
Figure~\ref{fig:masks} shows, instead, four cropped images after \gls*{prd} and their respective masks. It is noteworthy that most specular highlights are removed after \gls*{prd}.

At last, the \gls*{roi} is resized according to each approach proposed in Section~\ref{sec:proposed_approach}. The input sizes were chosen based on the original architectures of the approaches (see Table~\ref{tab:dimensions_overview}).

\begin{table}[!htb]
	\centering
	\caption{Image dimensions used in each approach.}
	\vspace{1mm}
	\label{tab:dimensions_overview}
	\begin{tabular}{@{}ccc@{}}
		\toprule
		\textbf{Approach} & \textbf{Image - Dimension} & \textbf{Mask - Dimension}  \\ \midrule
								
		\gls*{fcn}        & $ 320 \times 240 \times 3$ & $ 320 \times 240 \times 1$ \\
		\gls*{gan}        & $ 256 \times 256 \times 3$ & $ 256 \times 256 \times 3$ \\ 
		SegNet            & $ 320 \times 240 \times 3$ & $ 320 \times 240 \times 1$ \\						
		\bottomrule		
						
	\end{tabular}
\vspace*{-15pt}
\end{table}

\subsection{Evaluation Protocol}
\label{sub_sec:evaluation_protocol}

The performance evaluation of an automatic segmented mask is performed in a pixel-to-pixel comparison between the ground truth and the predicted image. Therefore, we use the following metrics: Precision, Recall and F-score. 

%


To perform a fair evaluation and comparison of the proposed approaches in all databases, we divided each into three subsets, being $40\%$ of the images for training, $40\%$ for testing and $20\%$ for validation. 
\section{Results and Discussions}
\label{sec:results}

The experiments were carried out using the protocol presented in Section~\ref{sub_sec:evaluation_protocol}. 
We also performed some additional experiments using the cross-sensor methodology.

\subsection{Proposed Protocol}

The results obtained by both the baseline (SegNet) and the proposed approaches are shown in Table~\ref{tab:results}. 
The baseline presented considerably worse results.
We believe this is due to the size of the training set, since SegNet was originally employed in a large dataset~\cite{radu2015robust}. Radu~et~al.~\cite{radu2015robust} generated a dataset with $54$,$000$ images using data augmentation. 
However, we did not have access to the database used by them, and thus a more direct comparison with their methodology was not possible to be done.

\vspace{-1mm}
\begin{table}[!htb]
	\centering
	\caption{Results achieved using the proposed protocol.}
	\vspace{1mm}
	\label{tab:results}
	\resizebox{\columnwidth}{!}{
		\begin{tabular}{ccccc}
			\toprule
			\textbf{Database} & \textbf{Approach} & \textbf{Recall \%} & \textbf{Precision \%} & \textbf{F-score \%} \\ \midrule
			\multirow{3}{*}{\acrshort*{ubirisv2}} & \gls*{gan} & $87.48$ $\pm$ $08.19$ & $87.10$ $\pm$ $08.16$ & $86.82$ $\pm$ $05.88$ \\
			& SegNet & $72.48 \pm 17.15$ & $87.52 \pm 08.53$ & $77.82 \pm 13.08$ \\
			& \textbf{\gls*{fcn}} & $\textbf{87.31}$ $\pm$ $\textbf{06.68}$ & $\textbf{88.45}$ $\pm$ $\textbf{06.98}$ & $\textbf{87.48}$ $\pm$ $\textbf{03.90}$ \\ \midrule
			\multirow{3}{*}{\gls*{miche}} & \gls*{gan} & $87.07 \pm 10.81$ & $86.39 \pm 12.02$ & $86.27 \pm 09.97$ \\
			& SegNet & $64.59 \pm 24.73$ & $83.39 \pm 18.53$ & $69.87 \pm 22.34$ \\
			& \textbf{\gls*{fcn}} & $\textbf{87.59}$ $\pm$ $\textbf{11.28}$ & $\textbf{89.90}$ $\pm$ $\textbf{09.82}$ & $\textbf{88.32}$ $\pm$ $\textbf{09.80}$ \\ \midrule
			\multirow{3}{*}{\acrshort*{gs4}} & \gls*{gan} & $85.72 \pm 12.53$ & $86.12 \pm 13.01$ & $85.20 \pm 11.31$ \\
			& SegNet & $66.50 \pm 26.34$ & $76.09 \pm 23.80$ & $67.92 \pm 23.87$ \\ 
			& \textbf{\gls*{fcn}} & $\textbf{88.24}$ $\pm$ $\textbf{12.03}$ & $\textbf{88.65}$ $\pm$ $\textbf{10.62}$ & $\textbf{88.12}$ $\pm$ $\textbf{10.56}$ \\ \midrule
			\multirow{3}{*}{\acrshort*{ip5}} & \gls*{gan} & $88.11 \pm 07.40$ & $87.71 \pm 07.71$ & $87.42 \pm 05.43$ \\
			& SegNet & $31.90 \pm 26.05$ & $79.40 \pm 32.93$ & $40.95 \pm 29.19$ \\ 
			& \textbf{\gls*{fcn}} & $\textbf{87.51}$ $\pm$ $\textbf{11.61}$ & $\textbf{89.32}$ $\pm$ $\textbf{05.22}$ & $\textbf{87.80}$ $\pm$ $\textbf{08.24}$ \\ \midrule
			\multirow{3}{*}{\acrshort*{gt2}} & \gls*{gan} & $86.20 \pm 15.02$ & $83.81 \pm 15.73$ & $84.50 \pm 14.28$ \\ 
			& SegNet & $73.77 \pm 21.20$ & $76.46 \pm 18.29$ & $72.33 \pm 18.26$ \\ 
			& \textbf{\gls*{fcn}} & $\textbf{87.86}$ $\pm$ $\textbf{12.23}$ & $\textbf{88.50}$ $\pm$ $\textbf{12.68}$ & $\textbf{87.94}$ $\pm$ $\textbf{11.59}$ \\ \bottomrule
		\end{tabular}}
		
\end{table}

Better results were obtained using the proposed approaches. In the \acrshort{ubirisv2} subset, the \gls*{gan}-based sclera segmentation attained a F-score value of $86.82$\%~$(\pm 5.88)$, while the approach based on \gls*{fcn} achieved $87.48$\%~$(\pm 3.90)$. Although there is little difference between the F-score values obtained by both methods, the standard deviation presented when using \gls*{fcn} was slightly lower than when \gls*{gan} was employed for the segmentation.

The same happened in all subsets used in our experiments, fact that makes us believe that the \gls*{fcn} approach is best suited for sclera segmentation. However, the results obtained with the \gls*{gan}-based segmentation should not be diminished, since they were very close to the best results.

\begin{figure}[!htb]
	\vspace{-5mm}
	\begin{center}
		\subfloat[][\gls*{gan} \label{fig:masksa}]{
			\includegraphics[width=0.31\columnwidth]{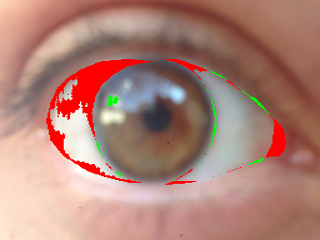}
		}
		\subfloat[][\gls*{fcn} \label{fig:masksb}]{
				\includegraphics[width=0.31\columnwidth]{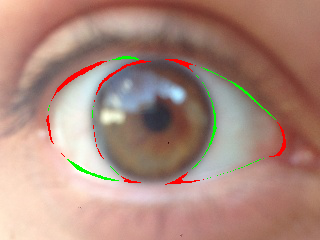}
		}
		\subfloat[][SegNet \label{fig:masksc}]{
				\includegraphics[width=0.31\columnwidth]{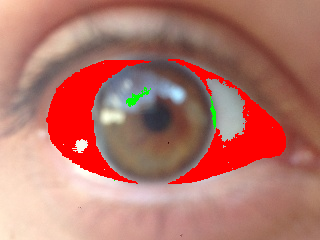}
		} \\ [-1ex]
		
		\subfloat[][\gls*{gan} \label{fig:masksd}]{
			\includegraphics[width=0.31\columnwidth]{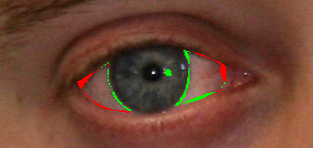}
		}
		\subfloat[][\gls*{fcn} \label{fig:maskse}]{
			\includegraphics[width=0.31\columnwidth]{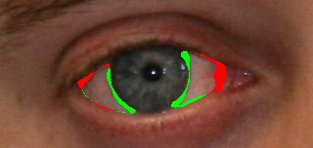}
		}
		\subfloat[][SegNet \label{fig:masksf}]{
			\includegraphics[width=0.31\columnwidth]{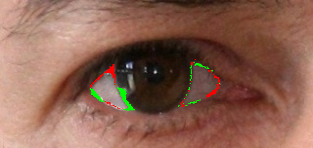}
		}
		
	\end{center}
	\caption{Samples of segmented scleras using the ground truth for highlighting errors: green and red pixels represent the FPs and FNs respectively.}
	\label{fig:masks3} 
	\vspace{-5mm}
\end{figure}

Here we perform a visual analysis. For this task, we randomly chose an image from the \acrshort*{ubirisv2} subset.
Figures~\ref{fig:masksa}, \ref{fig:masksb} and \ref{fig:masksc} demonstrate a very poor outcome in the segmentation of the sclera. As can be seen, the \gls*{fcn} approach presented a considerably better segmentation result when compared to the baseline and \gls*{gan}. The same occurs in many other images, but the results are not always so discrepant (see Figures~\ref{fig:masksd}, \ref{fig:maskse} and \ref{fig:masksf}). It is noteworthy the consistency presented with \gls*{fcn}-based segmentation technique, observed in all sclera images generated in this work.

\subsection{Additional Experiments Using Cross-Sensor}
\label{sub_sec:inter_sensor_result}

In this section, we present the results obtained using a cross-sensor methodology, where two experiments were performed. In the first one, we used \acrshort*{miche} (see Table~\ref{tab:overview_datasets}) as training set and \acrshort*{ubirisv2} as test set. In the second experiment, we inverted the order and used \acrshort*{ubirisv2} as training set and \acrshort*{miche} as test set.

Table~\ref{tab:miche} presents the results obtained using \acrshort*{miche} as training set and \acrshort*{ubirisv2} as test set. As we can see, the obtained F-score was very close to that obtained when the training and test sets were from the same database, reaching a F-score value $1$\% higher. However, in this case the best segmentation was achieved with the \gls*{gan}-based approach.

\begin{table}[!htb]
	\centering
	\caption{Results obtained using \acrshort*{miche} as training set and \acrshort*{ubirisv2} as test set.}
	\vspace{2mm}
	\label{tab:miche}
	
	\resizebox{\columnwidth}{!}{
		\begin{tabular}{cccc}
			\toprule
			\textbf{Approach} & \textbf{Recall \%}      & \textbf{Precision \%}   & \textbf{F-score \%}     \\ \midrule
			 \textbf{GAN}               & $\textbf{90.02}$ $\pm$ $\textbf{05.46}$  & $\textbf{85.96}$ $\pm$ $\textbf{07.90}$  & $\textbf{87.52}$ $\pm$ $\textbf{03.74}$  \\
			SegNet            & $83.98 \pm 08.85$           & $82.64 \pm 08.40$           & $82.58 \pm 05.35$           \\
			FCN               & $87.92 \pm 05.28$          & $87.04 \pm 07.11$           & $87.14 \pm 03.53$           \\ 
			 \bottomrule
		\end{tabular}
	}
	\vspace*{-10pt}

\end{table}

The same did not happen when we used \acrshort*{ubirisv2} as training set and \acrshort*{miche} as test set, as shown in Table~\ref{tab:ubiris}. The attained F-score values were much lower than those obtained when \acrshort*{miche} was used as training.

\begin{table}[!htb]
	\centering
	\caption{Results obtained using \acrshort*{ubirisv2} as training set and \acrshort*{miche} as test set.}
	\vspace{2mm}
	\label{tab:ubiris}
	\resizebox{\columnwidth}{!}{
		\begin{tabular}{ccccc}
			\toprule
			\textbf{Approach} & \textbf{Recall \%}      & \textbf{Precision \%}   & \textbf{F-score \%}     \\ \midrule
			 GAN               & $68.45 \pm 19.13$  & $71.05 \pm 21.57$  & $67.98 \pm 18.41$  \\
			SegNet            & $27.77 \pm 22.43$           & $60.58 \pm 37.42$           & $30.36 \pm 22.79$           \\
			\textbf{FCN}               & $\textbf{74.99} \pm \textbf{20.14}$          & $\textbf{77.41} \pm \textbf{16.15}$           & $\textbf{73.40} \pm \textbf{17.07}$           \\ 
			\bottomrule
		\end{tabular}
	}
	
	\vspace*{-12pt}
\end{table}

This might have occurred because the subset of the \acrshort*{miche} database used in this work has more diversity and it is larger than the \acrshort*{ubirisv2} subset, which allows the generated model to better discriminate the pixels belonging to~sclera.
\vspace{2pt}
\section{Conclusions and Future Work}
\label{conclusions}

This work introduced two new approaches for sclera segmentation and compared them with a baseline (SegNet) method chosen in the literature. 
Both proposed approaches (\gls*{fcn} and \gls*{gan}) attained higher precision and recall values in all evaluated scenarios. 
Furthermore, these approaches presented promising results when evaluated in cross-sensor scenarios.

We also manually labeled $1,300$ images for sclera segmentation. 
These masks will be available to the research community once this work is accepted for publication, assisting in the fair comparison among published works.

There is still room for improvements in sclera segmentation, so we intend to:
1) design new and better network architectures; 
2) create a unique architecture that integrates the detection stage of the periocular region; 
3) employ a post-processing stage to refine the segmentation given by the proposed approaches;
4) design a general and independent sensor approach, where firstly the image sensor is classified and then the sclera is segmented with a specific approach;
5) compare the proposed approaches with methods applied in other domains such as iris segmentation~\cite{7550055, lakra2018segdensenet} and periocular-based recognition~\cite{8101565}.


\section*{Acknowledgments}

The authors thank the National Council for Scientific and Technological Development~(CNPq) (\# 428333/2016-8 and \# 313423/2017-2) and the Coordination for the Improvement of Higher Education Personnel~(CAPES) for the financial support. We gratefully acknowledge the support of NVIDIA Corporation with the donation of the Titan Xp GPU used for this research.

\balance

{\small
\bibliographystyle{ieee}
\bibliography{bibtex}
}

\end{document}